\documentclass[pdflatex,sn-mathphys]{sn-jnl}

\jyear{2021}%

\theoremstyle{thmstyleone}%
\theoremstyle{thmstyletwo}%

\theoremstyle{thmstylethree}%

\begin{document}

\title[Article Title]{Attention Awareness Multiple Instance Neural Network}


\author*[1]{\fnm{Jingjun} \sur{Yi}}

\author*[1]{\fnm{Beichen} \sur{Zhou}}



\affil*[1]{\orgdiv{School of Remote Sensing and Information Engineering}, \orgname{Wuhan University}, \orgaddress{\street{Luoyu Road}, \city{Wuhan}, \postcode{430079}, \state{Hubei}, \country{China}}}




\abstract{Multiple instance learning is qualified for many pattern recognition tasks with weakly annotated data. The combination of artificial neural network and multiple instance learning offers an end-to-end solution and has been widely utilized. However, challenges remain in two-folds. Firstly, current MIL pooling operators are usually pre-defined and lack flexibility to mine key instances. Secondly, in current solutions, the bag-level representation can be inaccurate or inaccessible. To this end, we propose an attention awareness multiple instance neural network framework in this paper. It consists of an instance-level classifier, a trainable MIL pooling operator based on spatial attention and a bag-level classification layer. Exhaustive experiments on a series of pattern recognition tasks demonstrate that our framework outperforms many state-of-the-art MIL methods and validates the effectiveness of our proposed attention MIL pooling operators.}




\maketitle

\section{Introduction}\label{sec1}

\textbf{Multiple instance learning} (MIL) was firstly developed for drug prediction \cite{Dietterich1997Solving}. In MIL, an object is regarded as a bag and it contains some instances. Bag label is used for supervision but we do not have any instance labels. Each instance belongs to either positive instance or negative instance. If a bag contains at least one positive instance, then it should be categorized as positive. Otherwise, it is classified as negative. 
MIL mimics the real applications where we only have weakly annotated data. It was applied on a series of pattern recognition tasks such as object tracking \cite{Babenko2009Visual} and saliency detection \cite{Wang2013Saliency}.

\textbf{MIL and ANN.} Artificial neural network (ANN) is effective for machine learning. While combing other solutions with MIL the whole framework is often not end-to-end \cite{Andrews2003Support,wang2012discriminative,Maron1998Multiple}, the combination of MIL and ANN can be trainable and it is easier to adjust to different datasets and tasks \cite{Zhang2004Improve,Zhang2006Adapting,Wang2016Revisiting}.

Due to the strong feature representation capability, deep learning methods have developed rapidly in the past few years. Allowing for the fact that MIL is qualified for weakly annotated data, the combination of MIL of deep neural network is drawing increasing attention in the past few years \cite{Peng2017Multiple,Xu2015Multiple}.  

\textbf{Problem statement.} Although the combination of ANN and MIL offers a chance for end-to-end MIL solutions, however, several challenges still remain.
\begin{itemize}

\item {\em Limitation in mining key instances.}

Although some trainable MIL frameworks have been proposed, current approaches usually use pre-defined MIL pooling operators such as mean or maximum pooling \cite{Wang2016Revisiting,Yun2018Comparing}. As is clearly pointed out in \cite{Ilse2018Attention}, non-trainable MIL operators are not effective enough to find the key instances inside a bag and they lack the flexibility to adjust to different tasks and datasets.

\item {\em Gaps towards semantic representation.}

Current solutions to combine ANN and MIL belong to either embedding space paradigm or instance space paradigm. Under embedding space paradigm, MIL pooling operators can not offer a direct bag-level probability distribution \cite{Amores2013Multiple}. Under instance space paradigm, the inference of instance scores can be inaccurate because current non-trainable MIL pooling operators can be not qualified enough to mine the relation between instances and bags \cite{Ilse2018Attention}. This weakness effects negatively on the calculation of bag scores. 

\end{itemize}

\textbf{Motivation and contribution.} The objective of our work is two-folds.

\begin{itemize}

\item {\em Finding an approach to effectively mine instances relevant to the bag label.}
Bag score is calculated based on instance feature representation. As is stated above, pre-defined MIL pooling operators are not qualified enough to stress the key instances. With a solution further investigate the relation between instances and bags, bag representation capability can be enhanced.

\item {\em Enhancing the semantic representation capability for multiple instance neural networks.}
In current instance space paradigm, inaccurate instance inference leads to poor bag-level semantic representation, while in current embedding space paradigm there is no direct bag representation. Hence, the semantic representation ability remains to be improved. 
\end{itemize}

Our contribution can be summarized as follows.

(1)We propose a framework to combine MIL and neural networks. It is an end-to-end solution and can further mine the relation between bags and instances. Also, this framework is under the direct supervision of bag labels. The semantic representation capability is enhanced.

(2)We propose a trainable MIL pooling operation based on spatial attention mechanism. With attention awareness, different instances are assigned with different weights so that key instances can be stressed. Also, it directly outputs a bag-level classification distribution. 

(3)Our proposed attention awareness multiple instance neural network framework outperforms current MIL approaches on a series of pattern recognition tasks such as MIL classification, text classification and image classification. 

\section{Related Work}\label{sec2}
\textbf{MIL for aerial image.}
Vatsavai et.al. proposed a Gaussian multiple instance learning (GMIL) approach to describe the complicated spatial pattern in very high resolution remote sensing images and mapped the slums. This method outperformed many pix-level methods \cite{Vatsavai2013Gaussian}. More recently, Bi et.al. introduced multiple instance learning into aerial scene classification \cite{Bi2019Multiple}. The results reveal that MIL helps enhance the local semantic representation for aerial scenes.

\textbf{MIL for medical image.}
MIL has been widely applied in medical image processing. Similar to \cite{Vatsavai2013Gaussian}, Kandemir et.al. also utilizes Gaussian process in MIL for medical image classification \cite{Kumar2019Variational}. Hou et.al further investigated an approach to determine instance categories via a two-stage solution combining ANN and EM algorithms \cite{Hou2016Patch}, but it is still not an end-to-end solution. More recently, Ilse et.al. proposed an attention based MIL pooling and offered an end-to-end solution for multiple instance neural network \cite{Ilse2018Attention}.

\textbf{MIL for document classification.} 
Yan et.al. proposed a sparse MIL solution to build a robust structural representation for instances and bags \cite{Yan2016Sparse}. Zhou et.al. mined the relation inside instances for text classification \cite{Zhou2009Multi} by considering bags as graph and using features of nodes and edges to describe the relation between instances. More recently, Wang et.al. proposed a trainable multiple instance neural network (MINet) \cite{Wang2016Revisiting}. 

\textbf{Attention based MIL.}
To further mine the key instances inside a bag and to enhance the bag representation capability, the design of trainable MIL pooling operators has recently been reported. Attention models can assign different weights to different instances and can be adopted to develop a MIL pooling operator. Thus, Ilse et.al. developed a trainable MIL pooling operator based on gated attention mechanism for medical image classification. 

The major differences between our work and theirs include 1) In \cite{Ilse2018Attention}, it is under the embedding space and there is no direct bag representation, but our current work offers a direct bag probability distribution and is under the direct supervision of bag labels; 2) \cite{Ilse2018Attention} is based on gated attention  while our work is based on spatial attention; 3) We validate our solution on more backbones, more large-scale datasets and more pattern recognition tasks.

\section{Preliminary}\label{sec3}
\textbf{Classic MIL formulation.}
A bag with label $Y$ consists of a set of instances $X=\{x_1,x_2,\cdots,x_k\}$. Generally, each instance x$_i$ is labeled as either a positive instance (denoted as 1) or a negative instance (denoted as 0). Hence, the label $y_i$ of instance $x_i$ can be represented as $y_i =\{0,1\}$, here $i=0,1,\cdots,k$.

Traditionally, a bag is labeled as either a positive bag or a negative bag based on the following rule:

$$Y=
\begin{cases}
0 & \text{if $\sum_{i=1}^k{x_{i}}=0$}\\
1 & \text{else}
\end{cases} \eqno{(1)}$$

\textbf{Bag probability distribution assumptions.}
Classic MIL formulation can lead to gradient vanishing problems when training a multiple instance neural network. To this end, log-likelihood function is recently utilized for optimization \cite{Ilse2018Attention}. Here, bag label is distributed as a Bernoulli distribution with the parameter $\theta(X)\in[0,1]$.

\textbf{Decomposition of score function.}
Calculating the bag probability needs a score function $S(X)$ for the instance set $X$. Since MIL assumes that bag probability $\theta(X)$ must be permutation-invariant, as is stated in \cite{Zaheer2017Deep}, the score function $S(X)$ is a symmetric function and can be decomposed as follows.

$$ S(X)=g(\sum_{x \in X} f(X)) \eqno{(2)}$$

where $f$ and $g$ are transformations, which will be discussed later.

\textbf{Functions of decomposed transformations.}
Currently, the combination of ANN and MIL is under either instance or embedding space. 

For methods under instance space, here transformation $g$ is an identity transformation and instance-level classifier $f$ returns the score of each instance. Then, MIL pooling operator aggregates the instance scores and calculates the bag-level probability distribution $\theta(X)$. 

For methods under embedding space, here transformation $f$ transforms instances $X$ into embedding spaces. But the feature representation after MIL pooling needs further processing to generate $\theta(X)$.

Since our attention awareness multiple instance neural network belongs to the instance space paradigm where the bag probability distribution can be directly calculated, one major task is to develop a trainable MIL pooling operator to effectively mine key instances.

\section{Methodology}\label{sec4}
\subsection{Network Overview}

\begin{figure*}[h]
    \centering 
	\includegraphics[width=4.5in]{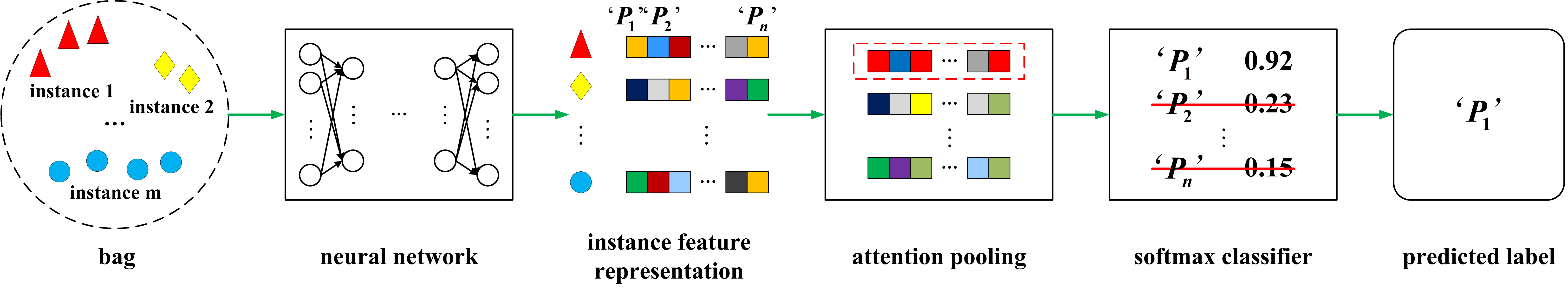} 
	\caption{Framework of our proposed attention awareness multiple instance neural network}  
	\label{fig2}   
\end{figure*}

As shown in Fig.~\ref{fig2}, our attention awareness multiple instance neural network consists of three parts, that is, an instance-level classifier, a trainable MIL pooling operator based on spatial attention, and a bag-level classification layer.

\subsection{Instance-level Classifier}
Under instance space paradigm, an instance-level classifier need to be built at first to generate the instance score or instance representation.

For two-category classification tasks, this instance representation $X$ can be quite simple. However, for multiple category classification tasks where there are multiple bag categories, this instance representation should have a same dimension corresponding to the bag category numbers so that the instance score on each bag category can be calculated. For example, we have $d$ bag categories and in a bag we have $n$ instances, then this instance representation $X$ should have a shape of $d\times n$.

\subsection{Attention Based MIL Pooling}

\textbf{Aggregation of instance prediction.}
The key function of MIL pooling operator under instance space paradigm is to aggregate the instance predictions into a bag representation.
To be specific, assume a bag contains $k$ instances and  instance $i$ corresponds to a prediction $\{p_{i}\}$, where $i=0,1,\cdots,k$. Let $p_{bag}$ denote the bag prediction and $O(\cdot)$ denote the aggregation function, then it can be represented as

 $$p_{bag}=O(\{p_{i}\}).\eqno{(3)}$$

\textbf{Attention module.}
To mine the key instances and to strengthen the bag representation capability, we introduce an attention model $H(\cdot)$ into MIL pooling so that different instances can be assigned to different weights.
To be specific, let $X$ denote the aforementioned instance-level feature representation, then each instance $x_i$ will correspond to an attention weight $a_{i}$ via

$$\{a_{i}\}=H(\{X\}).\eqno{(4)}$$

Our attention module consists of a fully connected layer and a sigmoid activation function. Let $W$ and $b$ denote the weight matrix and bias matrix of the fully connected layer, and let $sigmoid$ denote the aforementioned activation operation. Then, Equation 4 can be rewritten as

$$\{a_{i}\}=softmax(sigmoid(W X^T+b)),\eqno{(5)}$$

where $softmax$ denotes the softmax operation for normalization.  

\begin{figure}[h]
    \centering 
	\includegraphics[width=3.2in]{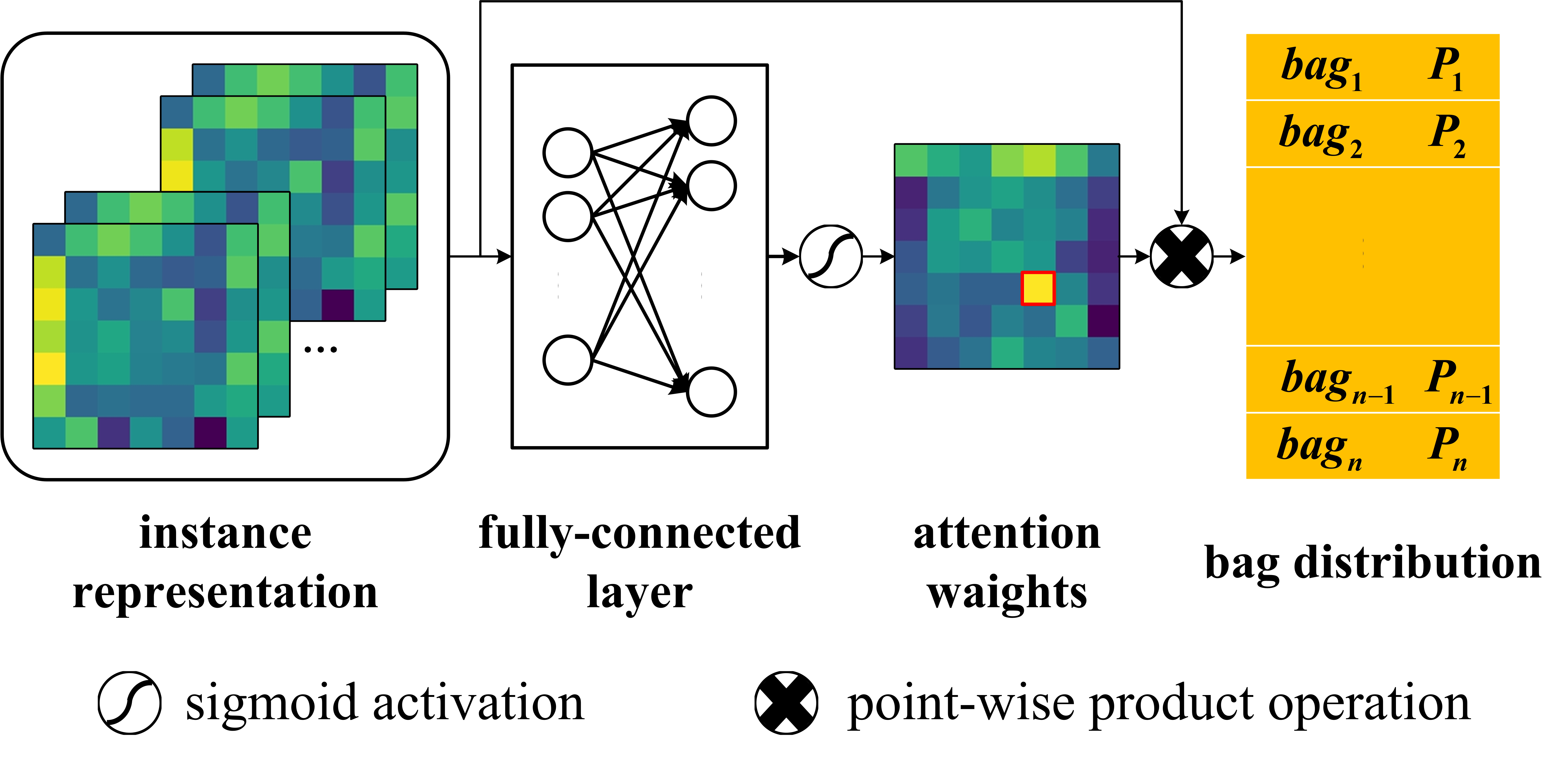} 
	\caption{Illustration of our attention based multiple instance pooling operator}  
	\label{fig3}   
\end{figure}

In Fig.~\ref{fig3}, our attention module is illustrated in detail.

\textbf{Weighted aggregation.} 
Current MIL pooling operators usually utilize pre-defined operators. To be specific, the function of $O(\cdot)$ is usually to select the mean value or maximum value of $\{p_{i}\}$.

In contrast, our attention based MIL pooling utilizes the weighted aggregation strategy. The $O(\cdot)$ is a convex combination of all the instance probability distribution vectors to generate the bag probability, represented as
 
 $$O(\{p_{i}\})=\sum \nolimits_i a_{i}p_{i}.\eqno{(6)}$$

Fig.~\ref{fig3} illustrates the situation when there are multiple bag categories.

\subsection{Bag-level Classification Layer}

With a bag-level probability distribution, we need to build a classification layer to form an end-to-end solution while keeping the whole network under the direct supervision of bag labels.

Here we choose the widely-used cross-entropy loss function to optimize the whole network and discuss the general situation when we have multiple bag categories. Let $N$ denote the number of bag categories, $y_c$ denote a bag's true label, and $\hat{y}_c$ denote the predicted bag value on category $c$. Then, this loss function can be represented as
$$L=-\frac{1}{N}\sum_{c=1}^n[y_c \log \hat{y}_c+(1-y_c)\log (1-\hat{y}_c)].\eqno{(7)}$$

\section{Experiment and Analysis}\label{sec5}
To investigate the effectiveness of our attention awareness multiple instance neural network on a series of pattern recognition tasks, we conduct the experiments on five standard classic MIL datasets, two aerial image datasets, one medical image dataset and three text classification datasets and compare it with other state-of-the-art methods.


\subsection{Classic MIL Datasets}
\textbf{Details.} 
Many former MIL methods have been validated on five standard MIL datasets (MUSK1, MUSK2, FOX, TIGER and ELEPHANT) \cite{Wang2016Revisiting}. Compared with the former solutions where ANN is combined with MIL in a end-to-end manner \cite{Wang2016Revisiting,Ilse2018Attention}, we keep backbone the same while inserting our attention module to test the performance of our solution. 

\textbf{Experimental setup.} 
The backbone we utilize is the mi-net \cite{Wang2016Revisiting,Ilse2018Attention}. The network structure is demonstrated in Table~\ref{tab1}. 
Hyper-parameters such as learning rate, iteration number and weight decay all keep the same with \cite{Wang2016Revisiting,Ilse2018Attention}.

\begin{table}[htp] 
    \centering
    \caption{Detailed network structure for classic MIL and MIL text classification}
    \begin{tabular}{ccc} 
    \hline
    Layer & mi-net & mi-net\_A(ours) \\
    \hline
    1 & fc-256+ReLU & fc-256+ReLU\\
    2 & dropout & dropout \\
    3 & fc-128+ReLU & fc-128+ReLU\\
    4 & dropout & dropout \\
    5 & fc-64+ReLU & fc-64+ReLU\\
    6 & dropout & dropout \\
    7 & fc-1+sigm & mil-attention \\
    8 & mil-mean/max & fc-1 \\
    \hline
    \end{tabular} 
    \label{tab1}
\end{table}

\textbf{Comparison.} 
Following the experiment protocols for standard MIL datasets, the mean accuracy and standard deviation of ten-folds cross validation are reported as the final result. The classification accuracy of our framework and other state-of-the-art MIL methods are listed in Table.~\ref{tab2}. 
Our solution (denoted as mi-net\_A) achieves the best classification accuracy on two out of these five datasets and comparable accuracy on the rest three datasets. More importantly, our solution outperforms end-to-end solutions mi-net, MI-Net and MI-Net with gated attention MIL pooling (denoted as MI-Net\_GA) on all these five datasets, indicating the effectiveness of our MIL pooling operation.

\begin{table}[htp]\tiny
    \centering
    \caption{Comparison of our framework and other state-of-the-art methods on classic MIL datasets.[1](\cite{Andrews2003Support}),[2](\cite{Zhou2009Multi}),[3](\cite{Wang2016Revisiting}),[4](\cite{Ilse2018Attention})}
    \begin{tabular}{cccccc} 
    \hline
    ~ & MUSK1 & MUSK2 & FOX & TIGER & ELEPHANT \\
    \hline
    mi-SVM [1] & 0.874$\pm$N/A & 0.836$\pm$N/A & 0.582$\pm$N/A & 0.784$\pm$N/A & 0.822$\pm$N/A \\
    MI-SVM [1] & 0.779$\pm$N/A & 0.843$\pm$N/A & 0.578$\pm$N/A & 0.840$\pm$N/A & 0.843$\pm$N/A \\
    mi-graph [2] & 0.889$\pm$0.033 & \textbf{0.903}$\pm$0.039 & 0.620$\pm$0.044 & \textbf{0.860}$\pm$0.037 & \textbf{0.869}$\pm$0.035 \\
    mi-net [3] & 0.889$\pm$0.039 & 0.858$\pm$0.049 & 0.613$\pm$0.035 & 0.824$\pm$0.034 & 0.858$\pm$0.037 \\
    MI-Net [3] & 0.887$\pm$0.041 & 0.859$\pm$0.046 & 0.622$\pm$0.038 & 0.830$\pm$0.032 & 0.862$\pm$0.034\\
    MI-Net\_GA [4] & 0.900$\pm$0.050 & 0.863$\pm$0.042 & 0.603$\pm$0.029 & 0.845$\pm$0.018 & 0.857$\pm$0.027 \\
    \textbf{mi-net\_A(Ours)} & \textbf{0.900}$\pm$0.063 & 0.870$\pm$0.048 & \textbf{0.630}$\pm$0.026 & 0.845$\pm$0.028 & 0.865$\pm$0.024 \\
    \hline
    \end{tabular} 
    \label{tab2}
\end{table}


\subsection{Aerial Image Dataset}
\textbf{Details.} Two aerial scene classification benchmarks, UCM \cite{Yi2013Geographic} and AID \cite{Xia2017AID}, are utilized to test the effectiveness of our framework. UCM dataset has 21 scene categories and each category has 100 samples. AID dataset has 30 scene categories with up to 10000 samples. 
In our MIL solution, the backbone is from commonly-utilized CNN models for aerial scene classification. By comparing our results to the results from these models, the improvement caused by MIL approaches can be revealed. Also, gated attention based MIL pooling \cite{Ilse2018Attention} on these backbones are tested to compare with our pooling operator.

\textbf{Experimental setup.} We utilize two widely-utilized CNN models in aerial scene classification as our backbone, that is, AlexNet and VGGNet-16. The first five convolutional layers of AlexNet and the first thirteen convolutional layers of VGGNet-16 serve as our instance-level classifier. Later on, these instance representations are fed into our attention based MIL pooling, which has 64 1$\times$1 convolutional layers and a sigmoid activation function. Finally, it is fed into a softmax classifier (demonstrated in Table.~\ref{tab3}).

\begin{table}[htp]
    \centering
    \caption{Detailed network structure for aerial image classification}
    \begin{tabular}{ccccc} 
    \hline
    Layer & Alex\_A(Ours1) & VGG\_A(Ours2) & AlexNet & VGGNet-16 \\
    \hline
    1 & conv11-96 & (conv3-64)$\times$2 & conv11-96 & (conv3-64)$\times$2 \\
    2 & maxpool & maxpool & maxpool & maxpool \\
    3 & conv5-96 & (conv3-128)$\times$2 & conv5-96 & (conv3-128)$\times$2 \\
    4 & maxpool & maxpool & maxpool & maxpool \\
    5 & conv3-384 & (conv3-256)$\times$2 & conv3-384 & (conv3-256)$\times$2 \\
    6 & conv3-384 & maxpool & conv3-384 & maxpool \\
    7 & conv3-256 & (conv3-512)$\times$2 & conv3-384 & (conv3-512)$\times$2 \\
    8 & conv1 & conv1 & ------ & maxpool \\
    9 & mil-attention & mil-attention & (fc)$\times$3 & (fc)$\times$3 \\
    10 & softmax & softmax & softmax & softmax \\
    \hline
    \end{tabular} 
    \label{tab3}
\end{table}

The backbone in both AlexNet and VGGNet-16 utilize parameters from pre-trained AlexNet or VGG-16 model on ImageNet as the initial parameters.

The model is trained by the Adam optimizer and has a batch size of 32. The initial learning rate is set to 0.0001 and 0.00005 in AlexNet based solution and VGG-16 based solution respectively. The learning rate is divided by 10 every 30 epochs. The training process does not terminate until 90 epochs are finished.
Moreover, to overcome the possible over-fitting problem, we use $L_2$ normalization with a parameter setting of $5\times10^{-4}$ and the dropout rate is set to be 0.2 in all solutions.

\textbf{Comparison.} Following the experiment protocols in \cite{Yi2013Geographic} and \cite{Xia2017AID}, the overall accuracy of ten independent runs are reported. All commonly-utilized training ratio settings (UCM 50\%, UCM 80\%, AID 20\% and AID 50\%) are all tested. Our results (denoted as Alex\_A and VGG\_A) are compared with the results from original AlexNet, original VGGNet-16, AlexNet with gated attention MIL pooling (denoted as Alex\_GA) and VGG with gated attention MIL pooling (denoted as VGG\_GA). All results are listed in Table.~\ref{tab4}.

It can be seen that our solution outperforms the backbone and the backbone with gated attention pooling operators under all four training ratio settings. Also, our solution outperforms the recently proposed multiple instance aerial scene classification solution in \cite{Bi2019Multiple}.

\begin{table}[htp]\footnotesize
    \centering
    \caption{Comparison of our framework and other state-of-the-art methods on aerial datasets.[1](\cite{Xia2017AID}),[2](\cite{Ilse2018Attention}),[3](\cite{Bi2019Multiple})}
    \begin{tabular}{ccccc} 
    \hline
    ~ & UCM50\% & UCM80\% & AID20\% & AID50\% \\
    \hline
    Alex [1] & 93.98$\pm$0.67 & 95.02$\pm$0.81 & 86.86$\pm$0.47 & 89.53$\pm$0.31\\
    Alex\_GA [2] & 96.02$\pm$0.44 & 97.64$\pm$0.64 & 90.60$\pm$0.31 & 92.91$\pm$0.37\\
    \textbf{Alex\_A(Ours1)} & \textbf{96.24}$\pm$0.42 & \textbf{98.22}$\pm$0.52 & \textbf{90.97}$\pm$0.14 & \textbf{93.71}$\pm$0.32\\
    VGG [1] & 94.14$\pm$0.69 & 95.21$\pm$1.20 & 86.86$\pm$0.47 & 89.53$\pm$0.31\\
    VGG\_GA [2] & 97.48$\pm$0.52 & 98.29$\pm$0.52 & 93.22$\pm$0.23 & 95.98$\pm$0.29\\
    \textbf{VGG\_A(Ours2)} & \textbf{97.92}$\pm$0.36 & \textbf{98.74}$\pm$0.48 & \textbf{93.99}$\pm$0.14 & \textbf{96.15}$\pm$0.29\\
    MIDCCNN [3] & 94.93$\pm$0.51 & 97.00$\pm$0.49 & 88.26$\pm$0.43 & 92.53$\pm$0.18\\
    \hline
    \end{tabular} 
    \label{tab4}
\end{table}

\subsection{Medical Image Dataset}
\textbf{Details.} 
Colon cancer dataset was published for medical image recognition \cite{Sirinukunwattana2016Locality}. It consists of 100 large-scale images with 22444 nuclei. Each nuclei belongs to one of the following four categories, that is, epithelial, inflammatory, fibroblast, and miscellaneous. Bags are split from the large-scale images and each bag is a 27$\times$27 image patch. 
We keep the same backbone in \cite{Sirinukunwattana2016Locality,Ilse2018Attention} for comparison.

\textbf{Experimental setup.} The backbone is softmax CNN in \cite{Sirinukunwattana2016Locality}. The convolutional features are later fed into our attention based MIL pooling. The detailed network structure is listed in Table.~\ref{tab5}. 
Hyper-parameters such as learning rate, iteration number and weight decay all keep the same with \cite{Sirinukunwattana2016Locality,Ilse2018Attention}.

\textbf{Comparison.} Following the experiment protocols in \cite{Sirinukunwattana2016Locality} and \cite{Ilse2018Attention}, the recall, precision and F-score of five independent runs are reported in Table.~\ref{tab6}.

It can be seen that our solution (denoted as softmaxCNN\_A) outperforms all other solutions relying on pre-defined MIL pooling operators (first four rows in Table.~\ref{tab6}). Also, it outperforms the recent proposed gated attention based MIL pooling operator (denoted as softmaxCNN\_GA) \cite{Ilse2018Attention}. Results on aerial and medical datasets indicate that our MIL pooling operator is qualified in these weakly-annotated visual tasks. 

\begin{table}[htp]
    \centering
    \caption{Detailed network structure for medical image classification}
    \begin{tabular}{ccc} 
    \hline
    Layer & softmaxCNN & ours \\
    \hline
    1 & conv4-36 & conv4-36 \\
    2 & maxpool & maxpool \\
    3 & conv3-48 & conv3-48 \\
    4 & maxpool & maxpool \\
    5 & (fc)$\times$3 & mil-attention \\
    6 & softmax classifier & softmax classifier \\
    \hline
    \end{tabular} 
    \label{tab5}
\end{table}

\begin{table}[htbp]
    \centering
    \caption{Comparison of our framework and other state-of-the-art methods on medical dataset.[1](\cite{Ilse2018Attention})}
    \begin{tabular}{cccc} 
    \hline
    ~ & precision & recall & F-score \\
    \hline
    instance+max [1] & 0.866$\pm$0.017 & 0.816$\pm$0.031 & 0.839$\pm$0.023\\
    instance+min [1] & 0.821$\pm$0.011 & 0.710$\pm$0.031 & 0.759$\pm$0.017 \\
    embedding+max [1] & 0.884$\pm$0.014 & 0.753$\pm$0.020 & 0.813$\pm$0.017 \\
    embedding+min [1] & 0.911$\pm$0.011 & 0.804$\pm$0.027 & 0.853$\pm$0.016 \\
    softmaxCNN\_GA [1] & 0.944$\pm$0.016 & 0.851$\pm$0.035 & 0.893$\pm$0.022 \\
    \textbf{softmaxCNN\_A(Ours)} & \textbf{0.950}$\pm$0.012 & \textbf{0.863}$\pm$0.020 & \textbf{0.904}$\pm$0.015 \\
    \hline
    \end{tabular} 
    \label{tab6}
\end{table}

\subsection{MIL Document Dataset}

\textbf{Details.} 
We validate the performance of our MIL framework on three benchmarks for MIL text classification utilized in \cite{Zhou2009Multi,Wang2016Revisiting}, that is, alt.atheism, comp.graphics and comp.windows.misc. The detailed information of these datasets are listed in \cite{Wang2016Revisiting}. Still, we keep the same backbone while inserting our attention module to test the performance of our solution. 

\textbf{Experimental setup.} Similar to the experiments in classic MIL datasets, the backbone we utilize is the mi-net \cite{Wang2016Revisiting,Ilse2018Attention}. Also, the attention based MIL pooling and the classification layer is the same with the experiments for classic MIL datasets (demonstrated in Table.~\ref{tab1}).
Hyper-parameters such as learning rate, iteration number and weight decay all keep the same with \cite{Wang2016Revisiting,Ilse2018Attention}.
\begin{table}[hpbt]\footnotesize  
    \centering
    \scriptsize
    \caption{Comparison of our framework and other state-of-the-art methods on MIL text datasets.[1](\cite{Zhou2009Multi}),[2](\cite{Wang2016Revisiting})}
    \begin{tabular}{cccc} 
    \hline
    ~ & alt.atheism & comp.graphic & comp.os \\
    \hline
    MI-Kernel [1] & 0.602$\pm$0.039 & 0.470$\pm$0.033 & 0.510$\pm$0.052 \\
    mi-Graph [1] & 0.655$\pm$0.040 & 0.778$\pm$0.016 & 0.631$\pm$0.015 \\
    mi-net [2] & 0.758$\pm$N/A & 0.830$\pm$N/A & 0.658$\pm$N/A \\
    MI-Net [2] & 0.776$\pm$N/A & 0.826$\pm$N/A & 0.678$\pm$N/A \\
    MI-Net\_DS [2] & \textbf{0.860}$\pm$N/A & 0.822$\pm$N/A & 0.716$\pm$N/A \\
    MI-Net\_RC [2] & 0.858$\pm$N/A & 0.828$\pm$N/A & \textbf{0.720}$\pm$N/A \\
    \textbf{mi-net\_A(Ours)} & 0.790$\pm$0.057 & \textbf{0.840}$\pm$0.052 & 0.710$\pm$0.057 \\
    \hline
    \end{tabular} 
    \label{tab7}
\end{table}

\textbf{Comparison.} Following the corresponding experiment protocols, the mean accuracy and standard deviation of ten-folds cross validation is reported as the final result. The classification accuracy of our and other state-of-the-art methods are listed in Table.~\ref{tab7}.
It can be seen that our solution (denoted as mi-net\_A)  achieves a comparable result when advanced strategies such as deep supervision (DS) and residual connection (RC) are utilized in mi-net \cite{Wang2016Revisiting}. 

\section{Conclusion}\label{sec6}

In this paper, we proposed an attention awareness multiple instance neural network framework. It consists of an instance-level classifier, a trainable MIL pooling operator and a bag-level classification layer. The trainable MIL pooling operator is based on spatial attention mechanism and it assigns different weights to different instances inside a bag and key instances can be stressed. Finally, the bag-level probabilities are fed into a classifier to form an end-to-end solution so that the multiple instance neural network is under the direct supervision of bag labels. 
Exhaustive experiments on a series of pattern recognition tasks including classic MIL problems, text classification, medical image classification and challenging aerial scene classification validate the effectiveness of our proposed framework.

\bibliography{sn-bibliography}


\end{document}